\pdfoutput=1

\documentclass[11pt]{article}

\usepackage[]{EMNLP2022}

\usepackage{times}
\usepackage{latexsym}

\usepackage[T1]{fontenc}

\usepackage[utf8]{inputenc}

\usepackage{microtype}

\usepackage{inconsolata}

\usepackage{graphicx}
\usepackage{amsmath}
\usepackage{multirow}
\usepackage{booktabs}

\definecolor{myred}{RGB}{227, 0, 0}
\definecolor{myblue}{RGB}{23, 76, 235}

\title{Cross-Modal Similarity-Based Curriculum Learning for Image Captioning}

\author{Hongkuan Zhang$^1$
\quad Saku Sugawara$^2$
\quad Akiko Aizawa$^2$ \\
\bf \quad Lei Zhou$^1$
\bf \quad Ryohei Sasano$^1$
\bf \quad Koichi Takeda$^1$\\
$^1$Nagoya University
\quad $^2$National Institute of Informatics\\
{\tt \{zhang.hongkuan.k5,zhou.lei.e1\}@s.mail.nagoya-u.ac.jp}\\
{\tt \{saku,aizawa\}@nii.ac.jp}
\quad {\tt \{sasano,takedasu\}@i.nagoya-u.ac.jp}\\
}

\begin{document}
\maketitle
\begin{abstract}
Image captioning models require the high-level generalization ability to describe the contents of various images in words. Most existing approaches treat the image--caption pairs equally in their training without considering the differences in their learning difficulties. Several image captioning approaches introduce curriculum learning methods that present training data with increasing levels of difficulty. However, their difficulty measurements are either based on domain-specific features or prior model training. In this paper, we propose a simple yet efficient difficulty measurement for image captioning using cross-modal similarity calculated by a pretrained vision--language model. Experiments on the COCO and Flickr30k datasets show that our proposed approach achieves superior performance and competitive convergence speed to baselines without requiring heuristics or incurring additional training costs. Moreover, the higher model performance on difficult examples and unseen data also demonstrates the generalization ability.

\end{abstract}

\section{Introduction}
\begin{figure}[t]
\centering
\includegraphics[width=0.5\textwidth]{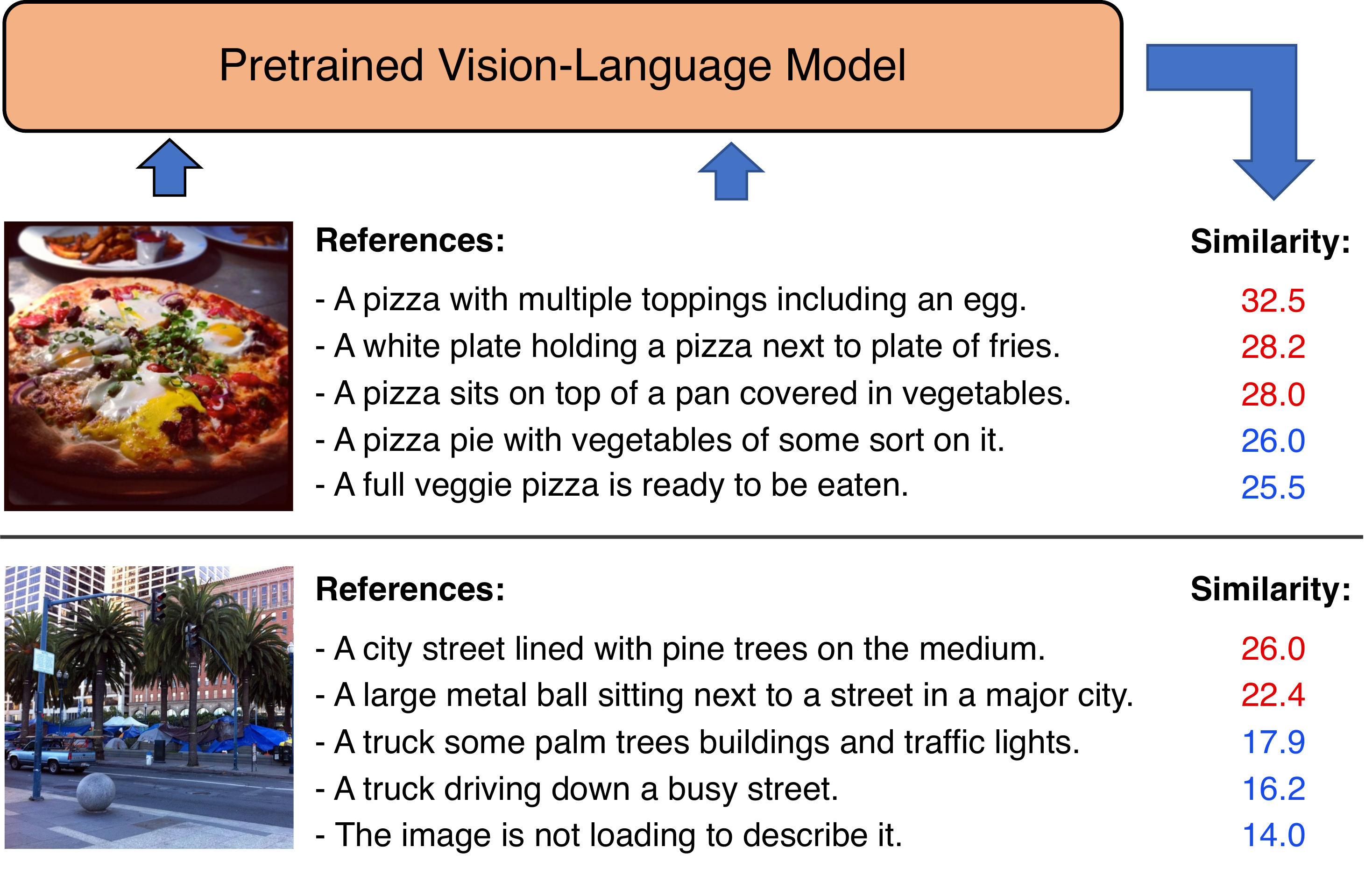}
\caption{Example of cross-modal similarity score for the caption data calculated by the pretrained VL model CLIP. Numbers with \textcolor{myred}{RED} and \textcolor{myblue}{BLUE} colors denote the higher and lower scores respectively.}
\label{caption-similarity-example}
\end{figure}

Image captioning has been widely investigated in computer vision and language research. However, most current methods treat image--caption pairs for training indistinctively, thus neglecting the difference in terms of learning difficulty. As illustrated in Figure \ref{caption-similarity-example}, an image is annotated with multiple references with diverse styles and complexity levels. Such diversity can introduce different levels of learning difficulty, and undertrained captioning models can be misled by wrong gradients when training on the difficult data \cite{dong2021dual}.


Curriculum learning (CL) has demonstrated improvements in model performance and training speed by presenting data sorted according to the learning difficulty \cite{bengio2009curriculum}. Existing image captioning approaches using CL have drawbacks in their difficulty measurements: 1) Requiring domain-specific knowledge or heuristics \cite{liu2021competence}; 2) Adding up the mono-modal difficulty scores without considering the cross-modal features \cite{alsharid2021course}; and 3) Requiring additional computational resources to train models on the target data (for the cases of bootstrapping methods) \cite{liu2021competence,dong2021dual}. 



We propose a simple yet efficient difficulty measurement using a pretrained vision--language (VL) model. Most VL models are pretrained with image--text matching tasks, which involve the calculation of the cross-modal similarity. The similarity reveals the model confidence in the image--text data relevance and a lower score indicates hard-to-determine or low-quality data \cite{lee-umic,hessel-etal}. As shown in Figure \ref{caption-similarity-example}, the VL model assigns higher scores to the highly relevant image--caption pairs usually with simple images and appropriate captions, while assigns lower scores to less relevant pairs often with complex images and low-quality captions. We consider the pairs with higher scores to be easier to learn and train the captioning model with training examples presented from easy to hard.

In our experiments on the COCO and Flickr30k datasets, models trained with our similarity-based CL achieve superior performance and convergence speed without domain-specific heuristics or additional pretraining cost. Moreover, using a VL pretrained model possessing the knowledge of the target data can further improve the performance. We also evaluate the trained model on difficult examples and unseen data, and the better performance of our method demonstrates the generalization ability. Last, our method brings higher improvement when applied to a smaller model, suggesting its applicability to scenarios in which fine-tuning large models is unfeasible.

\section{Related Work}
\paragraph{Curriculum Learning (CL)} CL is a method to train a model with sorted data to improve generalization and accelerate convergence. It has been explored in neural machine translation \cite{platanios-competence,liu-norm}, relation extraction \cite{huang2019self}, and natural language understanding \cite{xu2020curriculum} in the language field, or image classification \cite{wang2019dynamic,xiang2020learning} and semantic segmentation \cite{wei2016stc,huang2019self} in the vision field. We focus on the efficient data difficulty measurement of CL methods for image captioning.

\paragraph{Mono-Modal Difficulty Measurement} Difficulty measurement can be classified into predefined and automatic methods. Predefined methods require heuristics based on the data feature, such as the variety \cite{bengio2009curriculum} or number \cite{wei2016stc} of objects in an image and the length \cite{spitkovsky2010baby} or word rarity \cite{platanios-competence} in a sentence, to measure the data complexity for image classification and language generation. In contrast, automatic methods usually adopt a teacher model for difficulty measurement based on cross entropy \cite{weinshall2018curriculum,xu2020curriculum} or perplexity \cite{zhou2020uncertainty} 
to determine the model confidence and uncertainty.

\paragraph{Cross-Modal Difficulty Measurement} Difficulty measurement for caption data requires to consider the visual and language modalities. \citet{alsharid2021course} directly added visual and textual difficulty scores measured by Wasserstein distance and TF-IDF respectively for ultrasound image captioning. Similarly, \citet{liu2021competence} used both domain-specific heuristics and entropy from the bootstrapping model trained on the target data as difficulty measurements for medical report generation. In addition, \citet{dong2021dual} trained several bootstrapping models to evaluate generated captions with the BLEU score as the image difficulty. Unlike these works, we propose an efficient measurement based on cross-modal similarity to improve the model performance in the general domain.

\section{Methodology}
\subsection{Cross-modal Similarity}
To calculate the cross-modal similarity, we use either CLIP \cite{radford2021learning} pretrained on image--text pairs from the web or ViLT \cite{kim2021vilt} pretrained on labelled image-caption pairs for comparison. Specifically, given an image $X = (x_{1},...,x_{P})$ with $P$ patches and a text $Y = (y_{1},...,y_{T})$ with $T$ tokens, CLIP encodes each modality with individual encoder to obtain the visual feature $\mathbf{x}$ and textual feature $\mathbf{y}$ respectively. Then the similarity is calculated as:
\begin{equation}
  D_{\mathrm{CLIP\_sim}} = \cos(\mathbf{x},\mathbf{y}). \\
\end{equation}

While for ViLT, image and text inputs are concatenated with a prepended $\mathrm{[class]}$ token, and the inputs are encoded by a cross-modal encoder as:
\begin{equation}
  \mathrm{ViLT}(X,Y) = x'_{\mathrm{[class]}},x'_{1},...,x'_{P}, y'_{1},...,y'_{T}.\\
\end{equation}

The joint representation $x'_{\mathrm{[class]}}$ is then given to a pretrained fully-connected layer which is denoted as FFN to calculate the similarity as:
\begin{equation}
  D_{\mathrm{ViLT\_sim}} = \mathrm{sigmoid}(\mathrm{FFN}(x'_{\mathrm{[class]}})).\\
\end{equation}

\subsection{Training Schedule}
With the sorted dataset, we need to schedule when and how much harder data should be given during the training. Here we use the Baby Step learning \cite{spitkovsky2010baby} as our training schedule. The sorted dataset is equally divided into $L$ buckets, and the model is trained with the easiest bucket first. When the model performance on the validation set does not improve over several epochs, we consider the model has converged and then merge the harder bucket with current buckets to continue the training. Training terminates when all the buckets are used and the maximum number of training epochs is reached. In the experiments, we apply this training schedule to all the CL methods, and adjust the optimal number of buckets based on the model performance on the validation set. We use the notation Simi-CL for our proposed similarity-based CL. 

\subsection{Baseline Approaches}

\paragraph{Addup-CL} This method simply adds the difficulty scores of two modalities, and we use pretrained models to measure the difficulty score of each modality for a fair comparison. Specifically, we use pretrained object detector BUTD \cite{Anderson2017up-down} for visual difficulty $D_{v}$ and language model GPT-2 \cite{radford2019language} for textual difficulty $D_{t}$, and take the weighted sum to obtain the adding up difficulty $D_{\mathrm{addup}}$:
\begin{equation}
\begin{aligned}
D_{v} &= -\sum^{K}_{k=1} \sum^{N}_{n=1} p_{k,n} \log p_{k,n}, \\
D_{t} &= -\sum^{T}_{t=1}\log p(y_{t}|y_{<t}), \\
D_{\mathrm{addup}} &= \lambda \times D_{v} + (1-\lambda) \times D_{t}, \\
\end{aligned}
\end{equation}

\noindent where $K$ denotes the top-$K$ detected boxes with the highest confidence score, $N$ denotes the detected object classes, $p_{k,n}$ is the probability of the $n$-th class for the $k$-th box, and $\lambda$ denotes the weight.

\paragraph{Bootstrap-CL} This method requires training a model with target data in advance to provide the difficulty score. Specifically, we train the captioning model on each dataset with the regular strategy, then calculate the cross-entropy loss using the trained model as follows:
\begin{equation}
  D_{\mathrm{bootstrap}} = -\sum^{T}_{t=1} \log p(y_{t}|y_{<t},X).\end{equation}

\section{Experiments}

\subsection{Settings}
We performed experiments on the COCO and Flickr30k datasets and adopted the Karparthy splitting strategy \cite{karpathy2015deep}, obtaining 113k/5k/5k and 29k/1k/1k for the training/validation/test sets, respectively. We implemented the captioning model as a vanilla Transformer based on the publicly available codes \cite{luo2018discriminability}, and set the batch size to 10, learning rate to 3e-4, and dropout rate to 0.4 for all the experimental settings. 

For CL-related settings, the split numbers $L$ for the Baby Step learning were empirically determined to be 5 and 3 for COCO and Flickr30k, respectively. About the hyperparameters in Addup-CL, the number of the object detection classes $N$ is 1600 and we use the top-10 confident boxes to calculate the difficulty, and the weight $\lambda$ was set to 0.6 after the parameter tuning. For the similarity calculation, we used the base version of CLIP as the default model, and compare it with two versions of ViLT models which were fine-tuned on the COCO (\texttt{ViLT-CC}) and the Flickr30k (\texttt{ViLT-FL}) respectively by ViLT authors. We evaluated the performance with the COCO API and focused on four metrics: BLEU-4, METEOR, CIDEr, and SPICE.

\begin{table*}
\centering
\begin{center}
\begin{tabular}{lcccccccc}
\toprule
\multirow{2}{*}{Model} & \multicolumn{4}{c}{COCO} & \multicolumn{4}{c}{Flickr30k} \\
~ & B@4 & M & C & S & B@4 & M & C & S \\
\midrule
\textit{Transformer Baseline w/ pretraining} & & & & & & & & \\
LEMON \cite{hu2022scaling} & 40.3 & 30.2 & 133.3  & 23.3 & - & - & - & - \\\midrule
\textit{Transformer Baselines w/o pretraining} & & & & & & & & \\
VL-BART \cite{cho2021unifying} & 33.8 & 28.5 & 112.4  & 21.4 & - & - & - & - \\
Unified VLP \cite{zhou2020unified} & 35.5 & 28.2 & 114.3 &  21.0 & 27.6 & 20.9 & 56.8 & 15.3 \\
AoANet \cite{huang2019attention} & 37.2 & 28.4 & 119.8 & 21.3 & - & - & - & - \\
\midrule
\textit{Our Implemented Baselines} & & & & & & & & \\
Transformer & 35.7 & 27.9 & 113.0 & 20.9 & 27.7 & 21.8 & 58.5 & 16.0 \\
Transformer + Addup-CL & 35.2 & 27.9 & 114.2  & 21.0 & 26.5 & 21.5 & 56.6 & 16.0 \\
Transformer + Bootstrap-CL & 36.1 & 28.0 & 115.8  & 21.1 & 27.6 & 21.9 & 59.1 & 16.0 \\
\midrule
\textit{Our Proposed Methods} & & & & & & & & \\
Transformer + Simi-CL (ViLT) & 35.9 & 28.0 & 115.6 & 21.2 & 27.3 & 21.9 & 59.0 & 16.0
\\
Transformer + Simi-CL (CLIP) & 36.3 & 28.1 & 116.2 & 21.2 & 27.0 & \textbf{22.1} & 59.6 & 16.2
\\
Transformer + Simi-CL (ViLT-CC) & \textbf{36.4} & \textbf{28.2} & \textbf{117.1} & \textbf{21.4} & 27.5 & \textbf{22.1} & 61.0 & \textbf{16.3}
\\
Transformer + Simi-CL (ViLT-FL) & 36.0 & 28.0 & 115.9 & 21.0 & \textbf{28.5} & \textbf{22.1} & \textbf{61.8} & 16.2
\\
\bottomrule
\end{tabular}
\caption{Overall performance of CL-based methods and existing state-of-the-art models on COCO and Flickr30k. B@4, M, C, and S represent BLEU-4, METEOR, CIDEr, and SPICE, respectively.}
\label{main-table}
\end{center}
\vspace{-0.7em}
\end{table*}

\begin{figure}[t]
\includegraphics[width=\columnwidth]{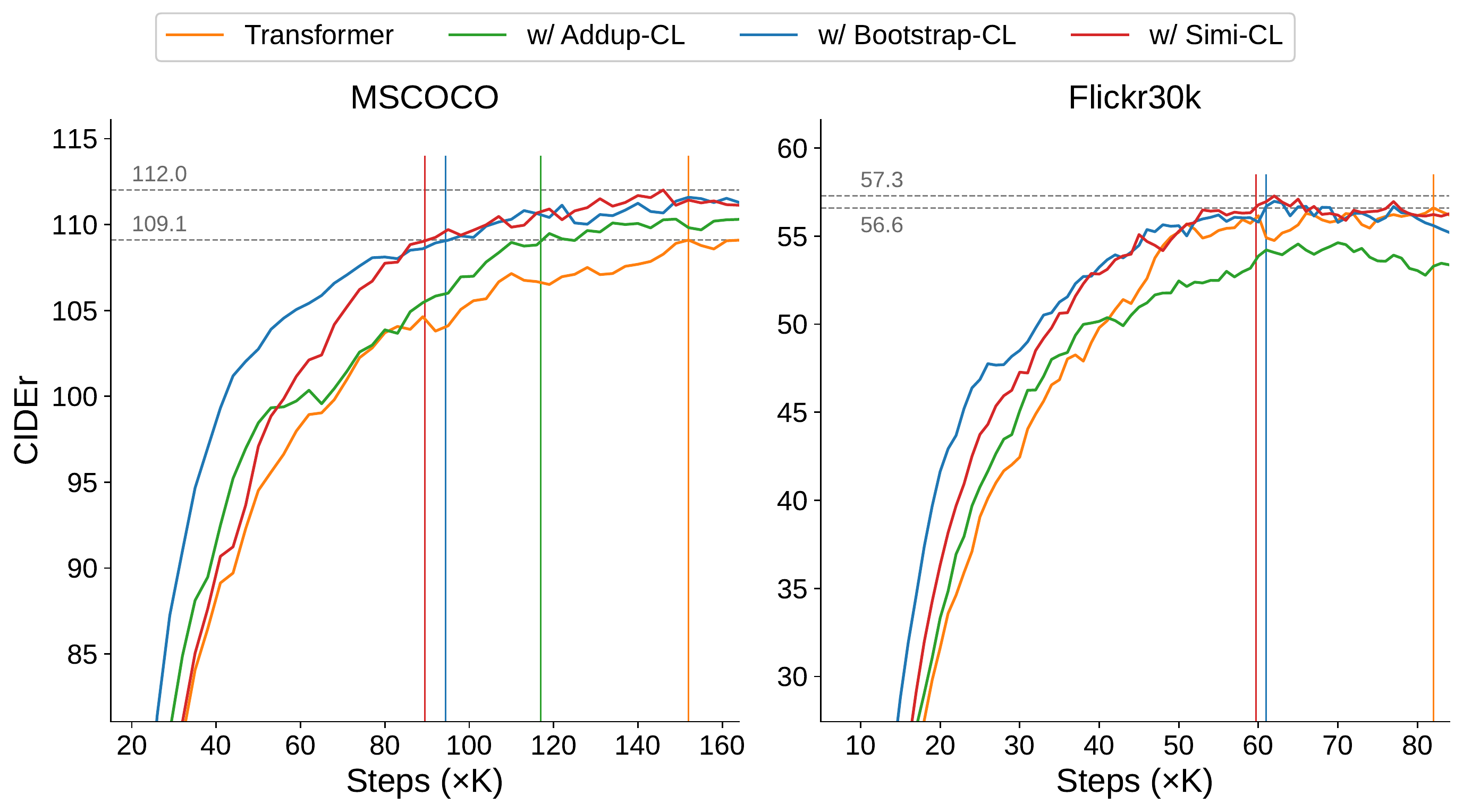}
\caption{Model performance variation on the validation sets of both datasets during the training.}

\label{valid-cider-score}
\end{figure}


\begin{table*}[ht]
\centering
\begin{tabular}{lcccccccccccccccc}
    \toprule
      & \multicolumn{8}{c}{Difficulty Level} \\
      \multirow{2}{*}{Model}  & \multicolumn{2}{c}{Level-1} & \multicolumn{2}{c}{Level-2} 
      & \multicolumn{2}{c}{Level-3} & \multicolumn{2}{c}{Level-4} \\
    \cmidrule(lr){2-3}
    \cmidrule(lr){4-5}
    \cmidrule(lr){6-7}
    \cmidrule(lr){8-9}
      ~ & B@4 & C & B@4 & C & B@4 & C & B@4 & C  \\
    \midrule
    Transformer & \textbf{71.0} & \textbf{199.5} & 36.5 & 120.9 & 4.5 & 84.0 & 0.7 & 47.8 \\
    + Bootstrap-CL & 57.1 & 172.0 & \textbf{37.8} & 122.5 & 27.0 & 97.8 & 18.3 & 71.2 \\
    + Simi-CL & 57.0 & 172.5 & 37.1 & \textbf{122.8} & \textbf{27.8} & \textbf{100.4} & \textbf{18.6} & \textbf{72.6} \\
    \bottomrule
\end{tabular}
\caption{Model performance on the divided COCO test sets with different difficulty levels.}
\label{model-performance-on-split-data}
\end{table*}

\subsection{Main Results}

The performance on the validation set is shown in Figure \ref{valid-cider-score}. For COCO, all the CL methods improve the performance, and accelerate the convergence speed towards the best vanilla model performance. Particularly, Simi-CL achieves better performance than Bootstrap-CL without additional training cost, and both methods outperform Addup-CL. For the Flickr30k dataset, we observe a similar phenomenon but with smaller improvement and Add-CL fails to improve the performance, which indicates that CL method is more efficient for the larger dataset. Since the vocabulary size of COCO is larger than Flickr30k (9,487 vs. 7,000), we suppose the difficulty measurement is efficacy for more diverse data, which requires further investigations.


For the model performance on the test sets listed in Table \ref{main-table}, the performance of CL-based models is consistent with that for the validation sets, achieving similar performance to existing Transformer baselines. Among the Simi-CL settings, using the ViLT model without fine-tuning can bring improvements similar to Bootstrap-CL but lower than the CLIP model that has a larger size. While using the ViLT models fine-tuned with in-domain and non-target data, the model achieves similar performance to CLIP with fewer parameters, and using ViLT models fine-tuned with the target data can further outperform CLIP, which reveals the efficacy of teacher models possessing the target data knowledge for better curricula design. More details about the measured difficulty score distributions can be found in Appendix \ref{dm distribution}. We also conduct the significance test for measuring the improvements which are described in Appendix \ref{significance-test}.


\subsection{Quantitative Analysis}

\paragraph{Performance on Divided Sets} To understand how CL contributes to the vanilla captioning model, we equally divide the COCO test set into four subsets based on the BLEU scores of captions generated by the vanilla model for the test images. The results in Table \ref{model-performance-on-split-data} show that the vanilla model performance is unbalanced among data with different difficulty levels, while both CL methods improve the performance on the harder subsets, and Simi-CL achieves the best performance.

\begin{table}[t]
\centering
\begin{tabular}{@{}lcccc@{}}
    \toprule
    Model & B@4 & M & C & S \\
    \midrule
    Transformer & 15.8 & 17.0 & 35.8 & 10.9 \\
    + Bootstrap-CL & 18.1 & 17.5 & 38.3 & 11.4 \\
    + Simi-CL & \textbf{18.6} & \textbf{18.2} & \textbf{39.8} & \textbf{11.7} \\
    \bottomrule
\end{tabular}
\caption{Cross-dataset performance evaluation using the best-performing COCO model for Flickr30k.}
\label{different-split-number}

\end{table}

\begin{figure*}[t]
  \centering
  \includegraphics[width=\textwidth]{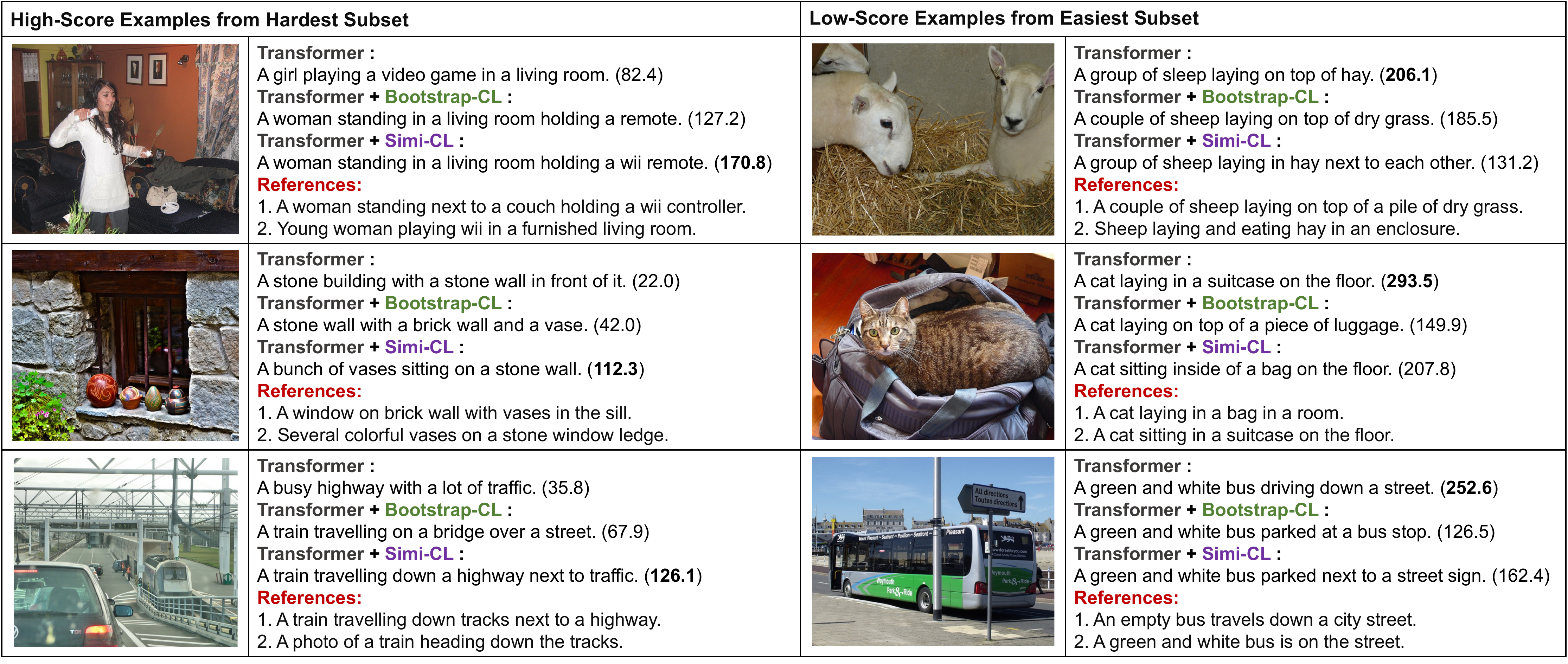}
  \caption{Samples of generated captions on divided sets. Number in the parentheses indicates the CIDEr score.}
  \label{qualatitive-analysis}
\end{figure*}

\begin{table}[ht]
\centering
\begin{tabular}{@{}lcccc@{}}
    \toprule
    Model & B@4 & M & C & S \\
    \midrule
    BUTD & 35.2 & 27.2 & 109.9 & 20.1 \\
    +Simi-CL & \textbf{36.2} & \textbf{27.8} & \textbf{113.0} & \textbf{20.6} \\
    \midrule
    AoANet & 36.8 & 28.0 & \textbf{117.2} & 21.3 \\
    +Simi-CL & \textbf{37.3} & \textbf{28.2} & 117.0 & \textbf{21.4} \\
    \bottomrule
\end{tabular}
\caption{Model performance on the COCO for applying Simi-CL to different model architectures.}
\vspace{-1em}
\label{different-architecture}
\end{table}

\paragraph{Model Generalization} To evaluate the model generalization ability, we test the model with cross-dataset evaluation referencing the former work \cite{torralba2011unbiased}. Specifically, we use the best-performing model trained with COCO to generate captions for the unseen test set from Flickr30k, obtaining the results listed in Table \ref{different-split-number}. The model performance maintains similar trends, with Simi-CL achieving the highest improvement and thus the best generalization ability.


\paragraph{Target Model Architecture} To investigate the effect of CL on different model architectures, we applied it to the LSTM-based model BUTD and more advanced Transformer-based model AoANet. The results are shown in Table \ref{different-architecture}. The improvement achieved by CL is higher when applied to a simpler architecture, which reveals the small-size model can benefit more from the large pretrained model.

\subsection{Qualitative Analysis}
We compare the captions from the vanilla model and CL-based models on the aforementioned divided subsets to understand the differences of generated captions as shown in Figure \ref{qualatitive-analysis}. On the hardest subset, we observed captions from Simi-CL can recognize objects more accurately such as \textit{wii remote} or \textit{vases} and describe contents in detail, which reveals the improved generalization ability. While on the easiest subset, we found that even if both CL-based models generate captions with similar or higher quality, low scores are given since the matched n-grams are less based on the limited references, which indicates the model-based metrics should be considered for the reference-free evaluation, and we leave it to our future work.



\section{Conclusion}
In this paper, we propose an efficient cross-modal similarity-based difficulty measurement for image captioning. Our proposed Simi-CL method boosts the model performance and training speed especially for larger datasets, and the pretrained models fine-tuned with target data can lead to further improvement. The improvement for data with different difficulty levels and data from other dataset indicates that Simi-CL achieves the highest model generalization ability. We also apply Simi-CL to different model architectures, and the higher improvement for the simpler model shows its practicality when only small-size models can be implemented in real-world scenarios.

\section*{Limitations}
The limitations of this paper are listed as follows:


\paragraph{Multiple Difficulty Measurements} We mainly focus on the CL method with a single measurement, but ensemble multiple measurements for model training may improve the model performance further or disturb each other, which requires further investigations. 


\paragraph{More Advanced Training Schedules} There are other advanced continuous CL training schedules such as the competence-based learning \cite{platanios-competence}, which samples the data from easy to hard gradually. We think our study is a baseline for follow-up work, and we believe a better training schedule will further boost the model performance.

\paragraph{Challengeable Datasets} There are several challengable image captioning datasets, such as the Novel Object Captioning (NoCaps) \cite{agrawal2019nocaps} and Conceptual Captions (CC) \cite{sharma2018conceptual}. Since the model trained with the CL method can handle the harder data with better generalization ability, we believe its performance on these datasets will be improved.


\section*{Acknowledgements}
We would like to thank the members of Aizawa Lab for their feedback on this work and the New Energy and Industrial Technology Development Organization (NEDO) for supporting this project.
\bibliography{emnlp2022}
\bibliographystyle{acl_natbib}

\appendix

\section{Difficulty Score Distribution}

\begin{figure}[ht]
\centering
\includegraphics[width=0.5\textwidth]{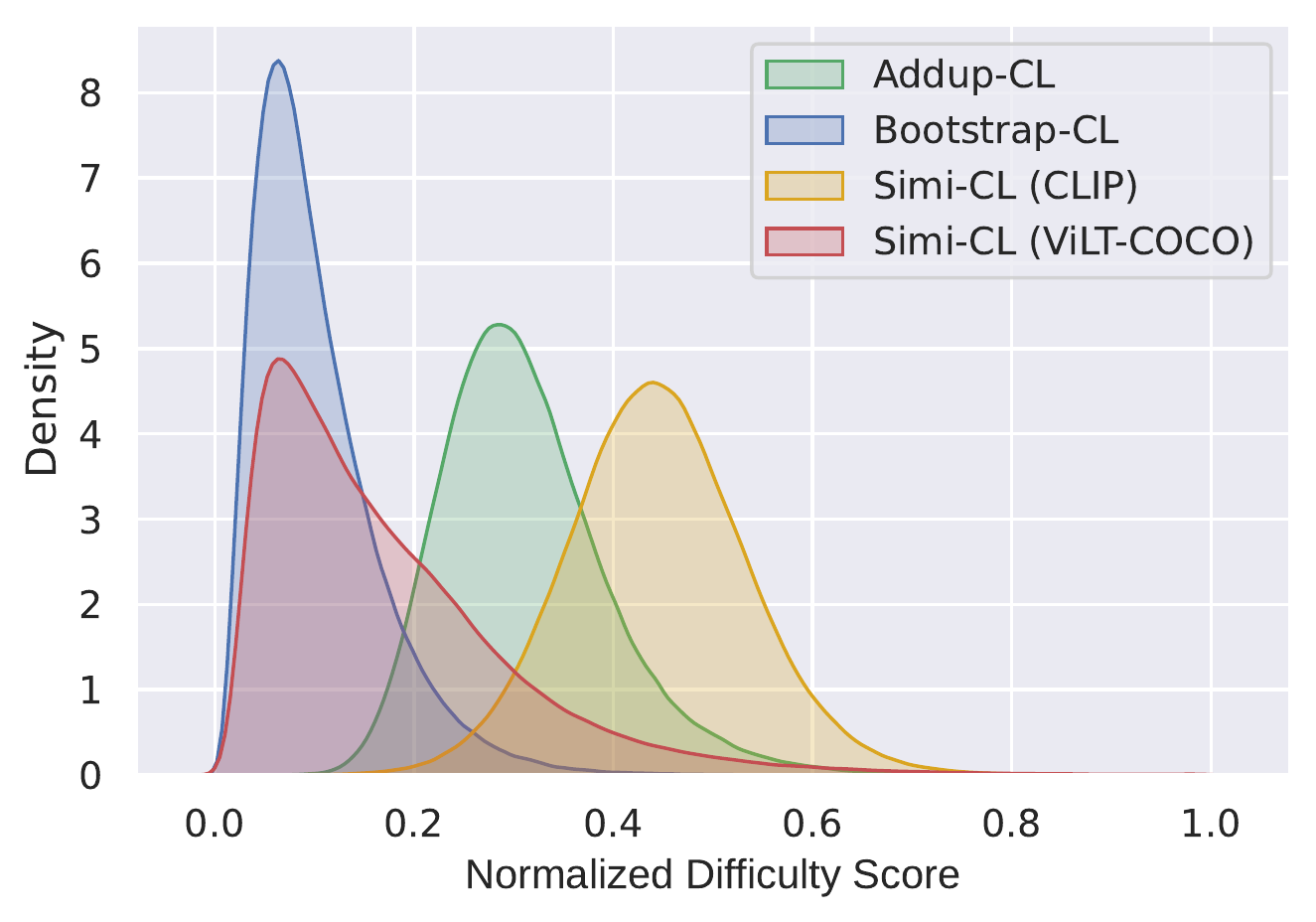}
\caption{Difficulty score distribution for the COCO training data under different difficulty measurements.}
\label{score-distribution}
\end{figure}

\label{dm distribution}
We provide distributions of normalized difficulty scores for the COCO training data under different difficulty measurements as shown in Figure \ref{score-distribution}. We find distributions of scores measured by both VL models have higher dispersion, which indicates VL models can differentiate the difficulty levels of the training data better, thus can achieve higher model performance. However, although the distribution of Addup method has higher dispersion than Bootstrap, the latter one brings higher improvement. We suppose that it is because a good measurement requires not only strong differentiation ability, but also the rational data sorting order, and Addup method cannot provide the appropriate sorting order by simply adding up the mono-modal scores for the difficulty measurement.

\section{Significance Test}
\label{significance-test}
We adopt the Bootstrap Test \cite{koehn2004statistical} that is widely used to compare two NMT systems' performance for our significance test. According to the sampling strategy in Bootstrap Test, we repeatedly sample images with replacements from the COCO test set to create 1000 sampled test sets (each contains 5000 images). Then we compute the metric scores for Bootstrap-CL and Simi-CL on all sampled sets. Finally we calculate the percentage of times that Simi-CL outperforms Bootstrap-CL to obtain the statistical significance as: Simi-CL is superior on BLEU-4, CIDEr, METEOR, and SPICE with p-value 0.032, 0.005, 0.001, and 0.001 respectively. If we set the significance level to 0.05, the improvements in all the metrics indicate the statistically significant improvement of our method.




\end{document}